\documentclass[letterpaper, 10 pt, conference]{ieeeconf}  
\IEEEoverridecommandlockouts 
\usepackage{cite}
\usepackage{amsmath,amssymb,amsfonts}
\usepackage{algorithmic}
\usepackage{graphicx}
\usepackage{textcomp}
\usepackage{xcolor}
\usepackage{graphics} 
\usepackage{epsfig} 
\usepackage{mathptmx} 
\usepackage{times} 
\usepackage{soul}   
\usepackage[pagebackref=true,breaklinks=true,colorlinks,bookmarks=false]{hyperref}
\usepackage{multirow}
\usepackage{multicol}
\usepackage{mathtools}
\usepackage{array}
\usepackage{bbding}

\title{\LARGE \bf Physical Human-Robot Interaction for Grasping in Augmented Reality via Rigid-Soft Robot Synergy}

\author{Huishi Huang$^{1,2,*}$, Jack Klusmann$^{1,3,*}$, Haozhe Wang$^{1,4,}$\textsuperscript{\textdagger}, Shuchen Ji$^{1}$, Fengkang Ying$^{1}$, Yiyuan Zhang$^{1}$, \\
John Nassour$^{3}$, Gordon Cheng$^{3}$, Daniela Rus$^{4,5}$, Jun Liu$^{2}$, Marcelo H. Ang Jr.$^{1}$, and Cecilia Laschi$^{1}$%
\thanks{*These authors contributed equally to this work.}%
\thanks{\textsuperscript{\textdagger}Corresponding author: wang\_haozhe@u.nus.edu}%
\thanks{$^{1}$Huishi Huang, Jack Klusmann, Haozhe Wang, Fengkang Ying, Shuchen Ji, Yiyuan Zhang, Marcelo H. Ang Jr., and Cecilia Laschi are with the Advanced Robotics Centre (ARC), National
University of Singapore (NUS), 117608, Singapore.}%
\thanks{$^{2}$Huishi Huang and Jun Liu are with the Institute of High Performance Computing (IHPC), Agency for Science, Technology and Research (A*STAR), 138632, Singapore.}%
\thanks{$^{3}$Jack Klusmann, John Nassour, and Gordon Cheng are with the Chair of Cognitive Systems, Technical University of Munich (TUM), 80333, Munich.}%
\thanks{$^{4}$Haozhe Wang and Daniela Rus are with the Singapore-MIT Alliance for Research and Technology (SMART) Centre, 138602, Singapore.}%
\thanks{$^{5}$Daniela Rus is with the Computer Science and Artificial Intelligence Lab (CSAIL), Massachusetts Institute of Technology, Cambridge,
MA 02139 USA.}%
}

\begin{document}

\maketitle
\thispagestyle{empty}
\pagestyle{empty}
\begin{abstract}
Hybrid rigid-soft robots combine the precision of rigid manipulators with the compliance and adaptability of soft arms, offering a promising approach for versatile grasping in unstructured environments. However, coordinating hybrid robots remains challenging, due to difficulties in modeling, perception, and cross-domain kinematics. In this work, we present a novel augmented reality (AR)–based physical human–robot interaction framework that enables direct teleoperation of a hybrid rigid–soft robot for simple reaching and grasping tasks. Using an AR headset, users can interact with a simulated model of the robotic system integrated into a general-purpose physics engine, which is superimposed on the real system, allowing simulated execution prior to real-world deployment. To ensure consistent behavior between the virtual and physical robots, we introduce a real-to-simulation parameter identification pipeline that leverages the inherent geometric properties of the soft robot, enabling accurate modeling of its static and dynamic behavior as well as the control system’s response.
\end{abstract}
\section{INTRODUCTION}

\begin{figure}[t]
\centering
    \includegraphics[width=\columnwidth]{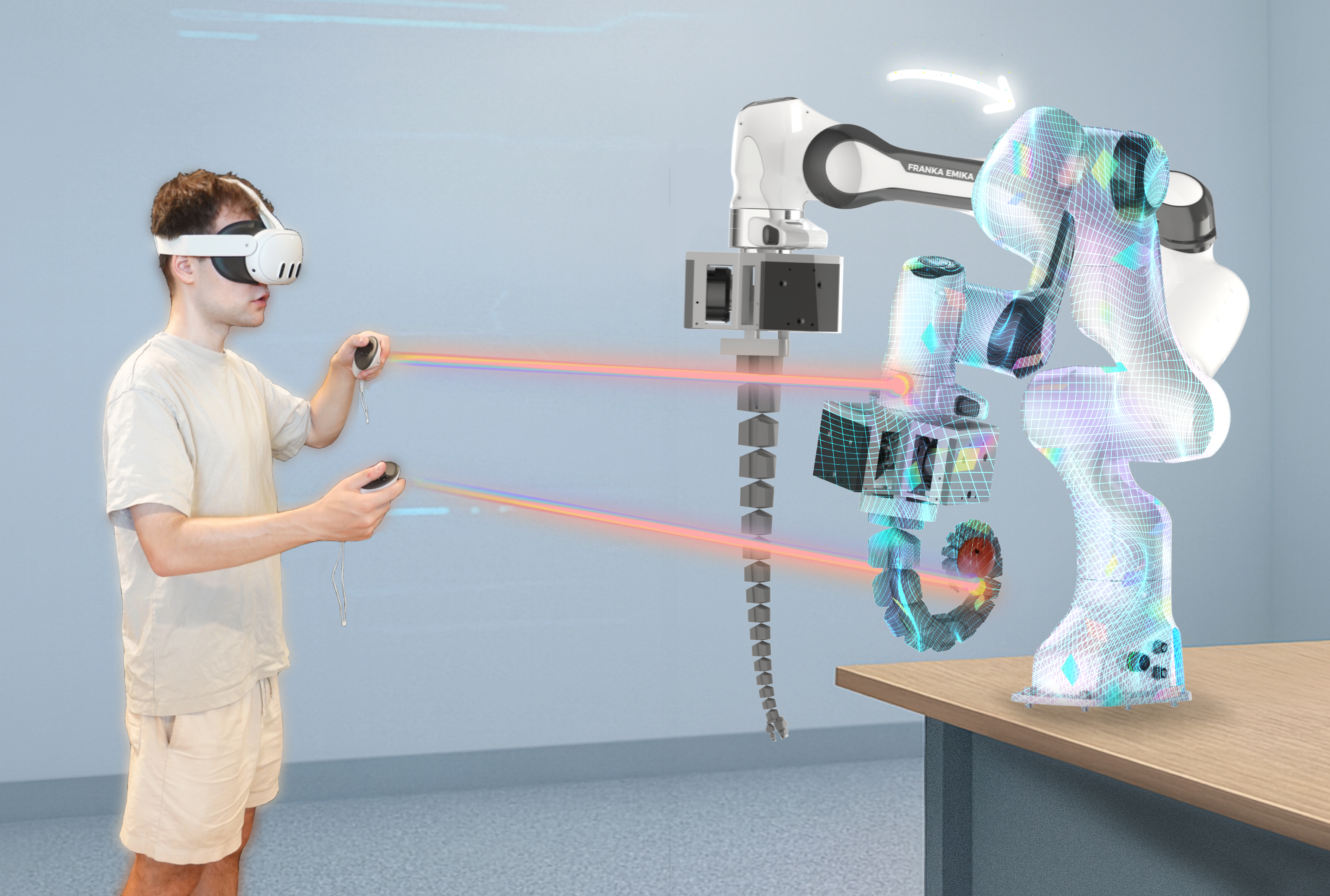}
\caption{Schematic diagram of the proposed AR interaction method.}
\label{ar_framework}
\vspace{-1.5mm}
\end{figure}

Teleoperation has long been a standard paradigm for robot control, particularly in industrial and medical applications. Conventional articulated robots are typically teleoperated through teaching pendants, joysticks, master-slave systems, or graphical user interfaces \cite{avgousti2016medical, dall2025towards}. 
Beyond these standard interfaces, recent advances have sought to enhance intuitiveness and immersion by multi-modality feedback, such as haptic feedback on wearable devices~\cite{thakur2024tetherless, uthai2025haptics}, augmented reality (AR)–based visual overlays on robot arms~\cite{pan2021augmented} or aerial robots \cite{walker2019robot}, and more intuitive body-machine interfaces~\cite{ajoudani2012tele}. 

However, soft continuum robots continue to pose significant challenges for teleoperation, and achieving reliable control remains an open research problem. Their high degrees of freedom (DoFs) and lack of analytical models complicate the mapping between human input and robot motion. Most existing efforts require labor-intensive manual tuning and rely on prototype-specific actuation strategies that limit generalizability. More advanced paradigms such as shared autonomy, user intention inference, or immersive extended reality-based teleoperation remain underexplored for soft robots due to difficulties in real-time modeling, sensing, and control.



Teleoperation of continuum arms can be generally categorized into end-effector control and shape control. End-effector control approaches allow users to specify a desired trajectory for the robot tip, for instance, through a digital pen interface~\cite{ouyang2018design}. Gesture-based methods have also been developed to map human motions to robot kinematics, enabling operations such as growth, retraction, steering, and rotation in growing robots~\cite{stroppa2020human}, or to execute constrained motions in two-segment continuum robots~\cite{lai2024gesture}. Shape control strategies are less common: some studies use shape correspondence to align the robot’s deformation with user-drawn curves for obstacle avoidance~\cite{ouyang2018design}, while others employ physical twin demonstrators to transfer desired configurations~\cite{guan2025control}.


Despite progress, relatively few studies have attempted to bridge hybrid rigid-soft structures with human-in-the-loop teleoperation. Prior work has typically treated the rigid and soft subsystems separately, controlling the rigid arm through kinesthetic teaching while actuating the soft arm using predefined motion patterns~\cite{huang2025grasping, montero2024mastering}. As such, the potential of hybrid teleoperation, which combines the precision of rigid mechanisms with the adaptability of soft structures, remains underexplored.

In this paper, we present a novel physical human-robot interaction (HRI) method to teleoperate a hybrid rigid-soft robotic system via AR (Fig. \ref{ar_framework}). Our hybrid system comprises a tendon-driven soft continuum arm mounted on a rigid robot arm, enabling high-DoF whole-arm grasping capabilities. We refer readers to our previous work in~\cite{huang2025grasping} for more details about the mechanical structure of our robotic system. Our HRI platform leverages the strengths of both subsystems, provides an immersive and straightforward teleoperation interface, and incorporates a virtual preview mechanism that allows users to evaluate and refine commands before execution on the physical robot. Through this approach, we demonstrate how AR can enable effective human-in-the-loop control of hybrid robots, thereby enabling safer and more versatile grasping in real-world environments.


To accurately superimpose the virtual robot onto the real one, a high-fidelity simulation is required to capture the robot’s static and dynamic behavior as well as its control system response. A common method to establish this reality-to-simulation (real-to-sim) transfer is via model formulation and parameter identification. Parameter identification is a well-established topic in rigid robotics~\cite{wu2010overview, khosla1985parameter}, frequently employed to determine quantities that are difficult to measure directly. However, parameter identification for soft robots remains underexplored due to several inherent challenges: 
\begin{enumerate}
    \item Soft robots possess a significantly larger parameter space than rigid robots.
    \item Soft robots lack inherent proprioception, requiring a more complex sensing mechanism to capture their deformation accurately.
    \item Soft robot modeling is computationally intensive, limiting real-time usability. 
\end{enumerate}

Analytical Lagrangian dynamics are built for relatively simple soft robot designs such as soft fingers~\cite{zhang2025novel} or soft grippers~\cite{wang2017soft}. More complicated hydrodynamics have also been taken into account for an aquatic soft robot~\cite{giorgio2017hybrid} for locomotion. Other studies adopt data-driven methods to identify both the dynamics and the controller parameters~\cite{wang2019parameter, bruder2019nonlinear}. 

Instead of deriving analytical models, we employ the physics-based simulation engine MuJoCo~\cite{todorov2012mujoco} to model the robot and identify simulation-oriented parameters using motion capture data and a numerical optimization algorithm. Our main contributions are summarized as follows:
\begin{enumerate}
\item We develop a novel physical human–robot interaction framework for a hybrid rigid–soft robot, enabling the user to specify end-effector positions for both the rigid and soft components within the shared workspace via an AR headset and controllers.
\item We establish a simulation-centered parameter identification framework for a cable-driven soft continuum arm, enabling the estimation of dynamic and control parameters within a general-purpose physics engine.
\item We validate the proposed interaction through a series of experiments demonstrating the system’s effectiveness for grasping.
\end{enumerate}

\section{METHODS}
In this section, we identify the selected methods to develop the HRI framework. Section \ref{me-sub1} presents the inverse kinematics approach and the reachability maps for the hybrid system. Section \ref{me-sub2} describes the modeling of our soft arm in MuJoCo. Section \ref{C} describes the parameter identification procedures for our soft robot. Section \ref{me-sub3} presents the methodology and practical applications of our HRI framework.

\subsection{Inverse Kinematics and Reachability Maps}\label{me-sub1}

\subsubsection{Rigid Robot}
We compute the inverse kinematics (IK) of the rigid robot numerically using an iterative solver to handle the redundancy of its seven degrees of freedom ~\cite{HeLiu2021}. 

Next, we generate a reachability map as a user input indicator for the following HRI interface. A dense sampling of feasible end-effector poses is performed via forward kinematics (FK), with the end-effector orientation constrained to point downward. While this restricted sampling does not capture the robot's full orientation dexterity, it aligns with practical AR teleoperation scenarios in which the end-effector predominantly points downward. The corresponding point cloud is enclosed to form a map as shown in Fig.~\ref{robot_workspace}(a).


\subsubsection{Soft Robot}
The spiral-shaped soft continuum arm poses a greater challenge for IK computation, which is essential for mapping the user's desired target input to the actuation value. To achieve a balance between accuracy and real-time deployability, we employ a lightweight neural network that directly learns the mapping from end-effector poses to tendon actuation lengths. The model is applied directly to both the simulated and physical robots.

Our model is a multilayer perceptron. The input to our model is an eight-dimensional vector comprising the gravity direction $\theta_G$, end-effector position $(x, y, z)$, and orientation represented as a unit quaternion $(q_w, q_x, q_y, q_z)$. Our model outputs the corresponding tendon lengths $(l_1, l_2, l_3)$ with three hidden layers, ReLU activations, and the Adam optimizer. 

We generate the training dataset from analytical FK simulations across multiple gravity directions using the model of \cite{huang2025grasping}, which consists of approximately $120,000$ samples uniformly distributed throughout the reachable workspace. We split the data into training and validation sets at a $80/20$ ratio. We export the trained network in ONNX format for integration into the AR-based control framework.


This architecture is sufficiently expressive to approximate the nonlinear relationship between end-effector pose and actuation while remaining compact enough for real-time inference on embedded hardware, enabling a low-latency IK controller that generalizes across the entire operating workspace.

\begin{figure}[htbp]
\centering
\includegraphics[width=\columnwidth]{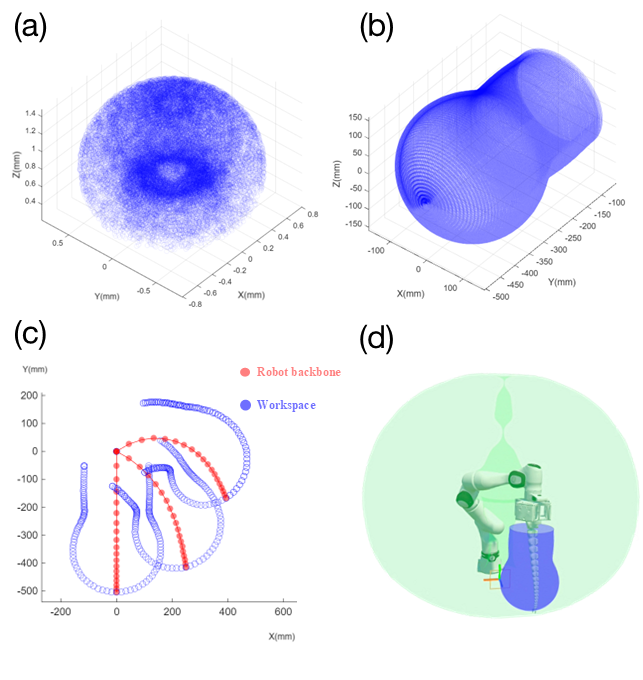} 
\caption{(a) Restricted workspace of the rigid robot. (b) Reachability map of the spiral-shaped soft arm with its backbone aligned with the gravity direction. (c) Reachability maps of our soft arm at $0^\circ$, $60^\circ$, and $120^\circ$ relative to the gravity direction. (d) Visualization of the restricted reachability map overlaid on the AR robot model.}
\label{robot_workspace}
\vspace{-2mm}
\end{figure}

We construct the reachability map for our soft arm by sampling forward kinematics (FK) outputs and enclosing the resulting point cloud within a 3D reachable surface, as shown in Fig. \ref{robot_workspace}(b). Since the effective reachability of our soft arm varies with gravity direction, which depends on the orientation of the rigid end-effector, we sample the FK outputs across multiple gravity directions---as shown in Fig. \ref{robot_workspace}(c)---to capture this dependence. We then update the reachability map in real time according to the current gravity direction. Fig. \ref{robot_workspace}(d) shows the restricted reachability map of our hybrid robotic system.

\subsection{Modeling as Pseudo-Rigid-Body}\label{me-sub2}

We develop a simulation model of our cable-driven continuum arm within the general-purpose physics engine MuJoCo. MuJoCo employs a pseudo-rigid-body model to represent our soft arm by discretizing it into multiple concatenated rigid segments, where each inter-segment connection is modeled as a torsional spring–damper system with two rotational degrees of freedom.

To understand which parameters need to be identified, we analyze the passive constitutive law for the pseudo-rigid-body model as follows. The total torque acting on each joint consists of the active driving torque (from the tendons) and the passive internal torque. Equation~\ref{eq1} specifically models the latter one, the viscoelastic resistance of the elastic joint, as a function of joint deformation and velocity. This internal torque acts as a counteracting force during cable-driven maneuvers. The mass and inertia of the joint are negligible compared to the link.

\begin{equation}\label{eq1}
    \tau = -K_{bending}\cdot(\theta-\theta_0)-D\cdot\dot{\theta}
\end{equation}
Here, the torque $\tau$ applied to each joint, characterized by bending stiffness $K_{bending}$ and damping coefficient $D$, drives the initial joint rotation $\theta_0$ toward the current rotation $\theta$ with angular velocity $\dot{\theta}$. The spring term generates a restoring torque that returns the joint to its equilibrium position, while the damping term suppresses rapid oscillations.


According to Euler-Bernoulli beam theory, the bending stiffness $K_{bending}$ can be expressed as Equation~\ref{eq2}:

\begin{equation}\label{eq2}
    K_{bending}=\frac{EI}{L}
\end{equation}

For a circular cross section with radius $r$, the area moment of inertia scales as $I \propto r^4$. The Young’s modulus $E$ is constant.

Since the robot geometry follows a logarithmic spiral constraint~\cite{wang2024spirobs}, the segment dimensions exhibit a fixed scaling ratio $\alpha$ along the backbone. This property allows the stiffness distribution to be initialized, where the stiffness $K_{bend,i}$ in segment $i$ is scaled along the arm according to Equation \ref{scale}:

\begin{equation}
\label{scale}
K_{bend,i} = K_0 \cdot \alpha^{(3* (i-1))}
\end{equation}
where $K_0$ is the stiffness of the base segment. Consequently, the bending stiffness scales as $K_{bending} \propto \alpha^3$.

The damping ratio $\zeta$ is a dimensionless parameter that characterizes the rate of energy dissipation in an oscillating system, as given by Equation~\ref{eq3}:

\begin{equation}\label{eq3}
   \zeta=\frac{D}{2\sqrt{K_{bending}\cdot\ m}}
\end{equation}

Here, $m$ denotes the mass of the segment. To initialize the damping parameters, each joint is approximated as a second-order vibrational system. When modeling our soft arm, we aim to maintain consistent vibration response behavior across all sections, which requires keeping the damping ratio $\zeta$ constant. Consequently, the damping coefficient scales as $D\propto\zeta{\cdot2\sqrt{K_{bending}\cdot m}=\alpha}^3$. 

 




\subsection{Parameter Identification}\label{C}

\begin{figure*}[htbp]
\centering
    \includegraphics[width=\textwidth]{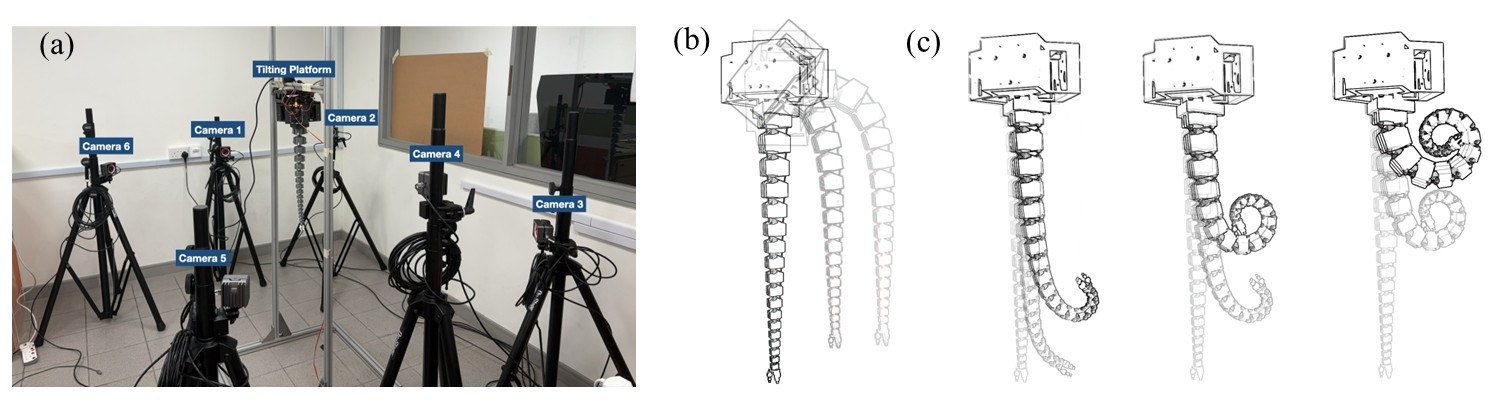}\\
\caption{(a). Experimental setup of our soft arm and the motion capture system with six infrared cameras. (b). Static tilting experiment to identify bending stiffness for each section, corresponds to Table~\ref{table1} Exp.1. (c) Curling and uncurling in different actuation statuses to identify damping properties for the robot and the actuator properties for the control system, corresponds to Table~\ref{table1} Exp.2 and 3.}
\label{experiment}
\end{figure*}

To accurately model the dynamic behavior and control system response of the soft continuum robot, we implement a staged parameter identification pipeline. The identification process is divided into three stages: (1) stiffness, (2) damping and friction, and (3) control gains and actuator force range. Each stage focuses on a specific subset of parameters, optimized using experimental data obtained from motion-capture trajectories recorded under varying conditions. The stages are performed sequentially, with each step building upon the previous one to ensure consistent dynamic behavior and control system response across the complete parameter space. We design a series of experiments for each stage (summarized in Table \ref{table1}), targeting static properties, dynamic behaviors, and the controller parameters. Our experiment setup is shown in Fig.~\ref{experiment}.

\begin{table*}[t]
\caption{Parameter Identification Experiments for our Soft Arm}
\renewcommand\arraystretch{1.5}
        \setlength\tabcolsep{6pt}
\label{table1}
\centering
\begin{tabular}{p{0.5cm}p{4.0cm}p{10.0cm}p{1.0cm}}
\hline
\textbf{Exp.} & \textbf{Objective} & \textbf{Implementation} & \textbf{Data} \\
\hline
\textbf{1} & Identify the joint stiffness for each section.  & 
Tilt the base by $0^\circ$, $30^\circ$, $60^\circ$, and $90^\circ$ relative to gravity and record the equilibrium section poses. & Static \\
\hline
\textbf{2} & Identify the joint damping for each section and tendon friction. & Release the robot from a fully bent state without actuation and record the oscillatory motion. & Dynamic \\
\hline
\textbf{3} & Identify actuator control gains and force range of the control system. & 
Perform curling and uncurling motions of the robot at actuation levels of $[20, 40, 60, 80, 100]$ mm, and record the resulting trajectories during each actuation cycle. & Dynamic \\
\hline
\end{tabular}
\vspace{-2.5mm}
\end{table*}



\subsubsection{Initialization}
Geometric parameters such as mass and inertia are initialized based on the material properties and robot geometry, while dynamic and control system parameters are estimated empirically.


Although the mass and inertia matrices can be directly derived from the robot's CAD model, real-world imperfections, such as non-uniform tendon routing, asymmetric 3D printing, and assembly inconsistencies, lead to discrepancies between theoretical and actual values. Since the primary objective is to capture the qualitative dynamic behavior of the continuum arm, mass and inertia are not explicitly re-identified. Instead, their effects are implicitly accounted for during the identification of bending stiffness and damping parameters.


\subsubsection{Stiffness Identification}




The first optimization stage focuses on the identification of joint stiffness coefficients  $K_{bending}$.
Simulations are conducted by tilting the robot to a fixed displacement and allowing it to relax under gravity, while recording the pose of body centroids using a motion-capture system. A Butterworth low-pass filter is subsequently applied to the recorded data to remove measurement noise. We minimize the following position-based loss function:

\begin{equation}
\begin{split}
\mathcal{L}
=
\frac{1}{N-1}
\sum_{i=2}^{N}
\mathrm{Huber}\!\left(
\frac{
d_{i}^{\text{sim}}
-
d_{i}^{\text{real}}
}{
\big|d_{i}^{\text{real}}\big|
+\varepsilon
},
\delta_{\text{pos}}
\right)
\end{split}
\label{eq:Ljoint_s}
\end{equation}


\noindent
Here, \(d_{i}^{\text{sim}}\) and \(d_{i}^{\text{real}}\) denote the Euclidean distances between the base segment and segment~\(i\) in the simulated and measured configurations, respectively. The loss compares the pairwise distances across all actuation sequences, focusing on the internal shape consistency of the robot rather than absolute coordinates. \(N\) is the number of robot segments, \(\varepsilon\) is a small constant for numerical stability, and \(\delta_{\text{pos}}\) is the Huber threshold controlling the transition between quadratic and linear penalization. The Huber loss ensures robustness to outliers, penalizing small residuals quadratically and large residuals linearly, thereby emphasizing accurate reproduction of the robot’s physical deformation.


The stiffness parameters are optimized in two stages: an initial coarse global identification optimized the stiffness of the first joint and scaled down for the following joints towards the tip, followed by a fine local adjustment of individual joint parameters within $\pm 10\%$ based on the coarse solution.


The optimization is performed using Differential Evolution (DE), a population-based, derivative-free global optimization algorithm. In DE, new candidate solutions are generated by combining scaled differences of randomly selected population vectors, followed by crossover and selection based on the objective value. For each candidate parameter set, a forward simulation is executed in MuJoCo and evaluated using the defined loss function.


\subsubsection{Damping}



Damping coefficients \(D_i\) are estimated using free-uncurling experiments, in which the robot is released from an actuated configuration and allowed to return to its neutral shape. To ensure that oscillatory behavior is accurately captured, 
a weighted loss \(\mathcal{L}_{\text{sum,t}}\) is employed, combining the positional loss \(\mathcal{L}_{\text{pos,t}}\) in Equation \ref{eq:Lpos_ts} and the velocity loss \(\mathcal{L}_{\text{vel,t}}\) in Equation \ref{eq:Lvel_ts}. The second term penalizes discrepancies in the dominant oscillations and decay behavior, thereby improving the identification of viscoelastic effects that are only weakly observable through a purely geometric objective.

\begin{equation}
\begin{split}
\mathcal{L}_{\text{pos},t}
&=
\frac{1}{N-1}
\sum_{i=2}^{N}
\mathrm{Huber}\!\left(
\frac{
d_{i}^{\text{sim}}(t)
-
d_{i}^{\text{real}}(t)
}{
\big|d_{i}^{\text{real}}(t)\big|
+\varepsilon
},
\delta_{\text{pos}}
\right),
\end{split}
\label{eq:Lpos_ts}
\end{equation}

\begin{equation}
\begin{split}
\mathcal{L}_{\text{vel},t}
&=
\frac{1}{N-1}
\sum_{i=2}^{N}
\mathrm{Huber}\!\left(
\frac{
\Delta d_{i}^{\text{sim}}(t)
-
\Delta d_{i}^{\text{real}}(t)
}{
\big|\Delta d_{i}^{\text{real}}(t)\big|
+\varepsilon
},
\delta_{\text{vel}}
\right),
\end{split}
\label{eq:Lvel_ts}
\end{equation}

\begin{equation}
\mathcal{L}_{\text{sum},t}
=
w_{\text{pos}}\,\mathcal{L}_{\text{pos},t}
+
w_{\text{vel}}\,\mathcal{L}_{\text{vel},t}.
\label{eq:Ljoint_ts}
\end{equation}


\noindent
Here, \(d_{i}^{\text{sim}}(t)\) and \(d_{i}^{\text{real}}(t)\) denote the Euclidean distances between the base segment and segment~\(i\) at time step~\(t\) in the simulated and measured trajectories, respectively. \(\Delta d_{i}(t)\) represents the temporal difference of these distances, capturing segment-wise velocity behavior. The time index \(t\) spans all frames within a sequence, \(t \in [1, T_s]\), with \(\mathcal{L}_{\text{vel},1}=0\). \(\varepsilon\) is a small constant for numerical stability, while \(\delta_{\text{pos}}\) and \(\delta_{\text{vel}}\) are Huber thresholds. At each time step, the position and velocity terms are weighted by \(w_{\text{pos}}\) and \(w_{\text{vel}}\) to form \(\mathcal{L}_{\text{sum},t}\) in Equation \ref{eq:Ljoint_ts}. These timestep losses are then averaged over each sequence and subsequently across all actuation trials to obtain the final loss \(\mathcal{L}_{\text{sum}}\).

Coarse damping factors are first optimized jointly across all segments, followed by fine-tuning of individual segment parameters within \(\pm10\%\) of the coarse solution.



\subsubsection{Control System Parameters}

The control behavior of the system is simplified and implicitly modeled as a position actuator governed by a proportional–derivative (PD) controller. After the stiffness and damping parameters are determined, the control-related parameters, namely the proportional gain \(k_p\), velocity gain \(k_v\), force range \(F_{\text{range}}\), and actuator time constant \(\tau_m\), are optimized. Assuming critical damping, the relationship between the proportional and derivative gains is initialized as:

\begin{equation}
k_v = 2 \sqrt{k_p \, m}
\end{equation}

The same trajectory-based loss function, \(\mathcal{L}_{\text{sum,t}}\), is applied across all time frames to evaluate the overall shape convergence throughout each motion.

\subsection{Physical Human Robot Interaction}\label{me-sub3}


In the proposed framework, the user operates two AR-tracked joysticks equipped with virtual laser pointers. As illustrated in Fig.~\ref{ar_framework}, the endpoint of one pointer specifies the desired position of the rigid arm end-effector, while the other defines the target location of the soft arm tip. The laser pointers are retractable, enabling intuitive 3D spatial control.

Before executing the motion on the physical robot, a virtual replica of the robot performs the commanded action in simulation, including the grasping attempt. This preview mechanism allows the user to assess whether the command aligns with the intended goal and to make adjustments through iterative trials in the virtual environment. Once the user confirms the command, the motion is executed on the physical robot, reducing trial-and-error during real hardware interaction and enhancing efficiency, safety, and user confidence.

\begin{figure}[htbp]
\centering

\centering
    \includegraphics[width=\columnwidth]{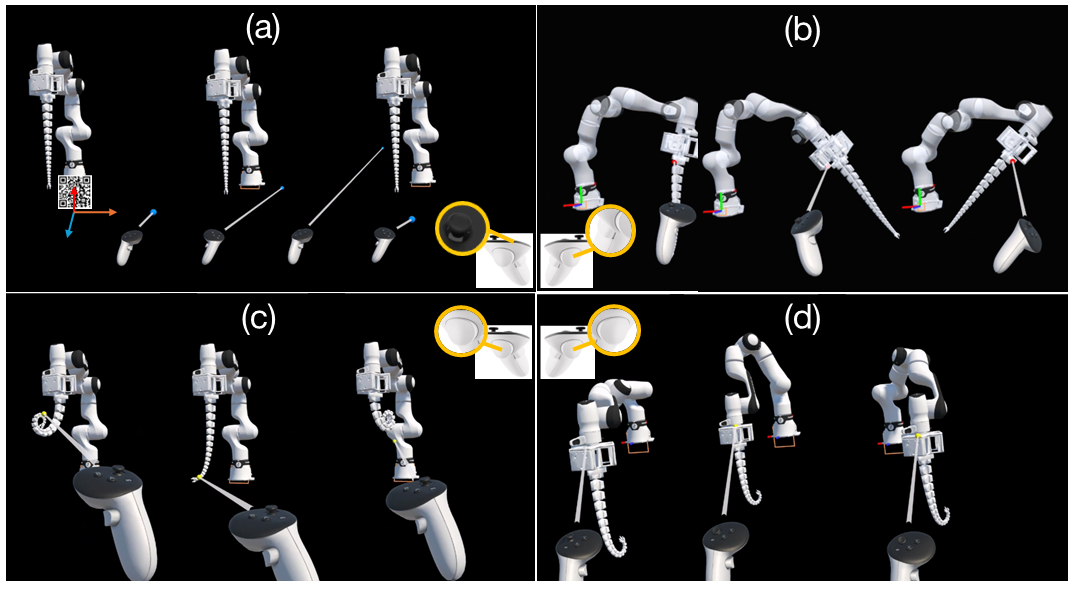} \\
    \includegraphics[width=\columnwidth]{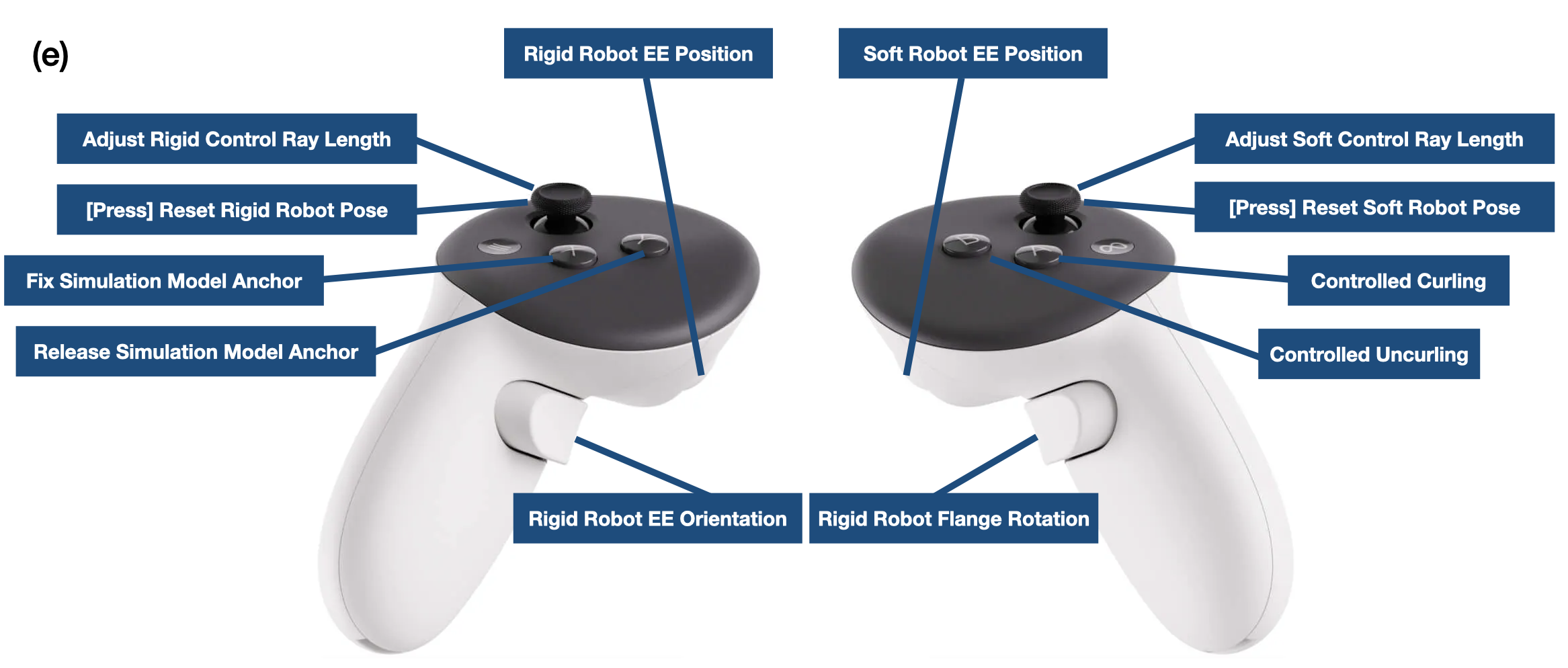}
\caption{Additional details are provided in the supplementary video. (a) Adjustment of the ray length projected from the joystick; (b) Tilting the joystick to change the orientation of the rigid robot for alignment with the joystick. (c) Control of the soft robot’s curling and uncurling motions by specifying the end-effector position using the ray. (d) Control of the rigid robot’s end-effector position using the ray. (e) Overview of the joystick functions.}
\label{ar_control_method}
\vspace{-2.5mm}
\end{figure}


The trunk-like robot is constrained to bending motions only. This enables a unique shape mapping under single-DoF actuation, which is sufficient for fundamental interaction and grasping tasks. Specifically, planar bending can be achieved through either single-cable dorsal actuation or symmetric dual-cable ventral actuation, resulting in a smooth spiral configuration that has been demonstrated to successfully grasp hanging objects~\cite{huang2025grasping}. When the user specifies a position outside these planes, the rigid arm end-effector automatically rotates to reorient the soft arm so that its bending plane aligns with the user-defined direction in the virtual space.




\section{EXPERIMENTS AND RESULTS}
We first present the hardware and experimental setup (Section \ref{ex-sub1}), then demonstrate the performance of the real-to-sim parameter identification (Section \ref{ex-sub2}), and finally describe the grasping experiments (Section \ref{ex-sub3}).

\subsection{Hardware and Experimental Setup}\label{ex-sub1}


The soft robot used in this experiment consists of a three-cable-driven, spiral-based soft arm connected to a tiltable base mount at one end and an end-effector gripper at the other. The physical prototype is fabricated via 3D printing using thermoplastic polyurethane (TPU-95A) with a $100\%$ infill rate to minimize structural inconsistencies. During data collection and user experiments, cable lengths are actuated by three independent motors (DJI GM6020 brushless DC motors). An OptiTrack motion capture system with six cameras (Fig. \ref{experiment}) tracks the real-time poses of 18 key segments at 120 Hz, enabling reliable reconstruction of the arm’s 3D backbone trajectory. This setup is used to conduct the parameter identification experiments summarized in Table~\ref{table1}.

We use the optimized parameters obtained in Section~\ref{C} to calibrate the soft robot model in MuJoCo 3.3.1. The calibrated MuJoCo XML model is subsequently imported into Unity 6 using the official MuJoCo-for-Unity plugin, which enables the MuJoCo physics engine to run natively within Unity. In this setup, MuJoCo remains responsible for all physics simulation, while Unity is used exclusively for rendering, AR visualization, and user interaction. The system is deployed on a Meta Quest 3 operating at 60 Hz.

\subsection{Parameter Identification}\label{ex-sub2}





\begin{figure}[htbp]
\centering
    \includegraphics[width=\columnwidth]{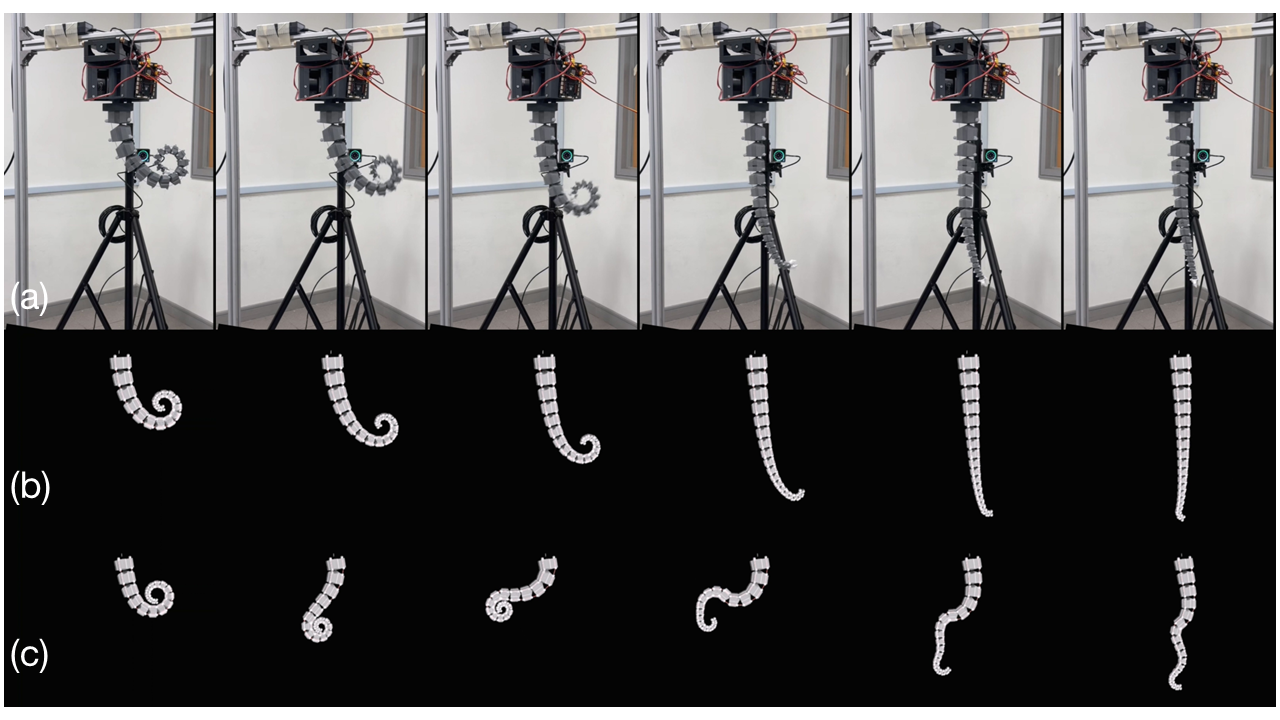} \\
    \includegraphics[width=\columnwidth]{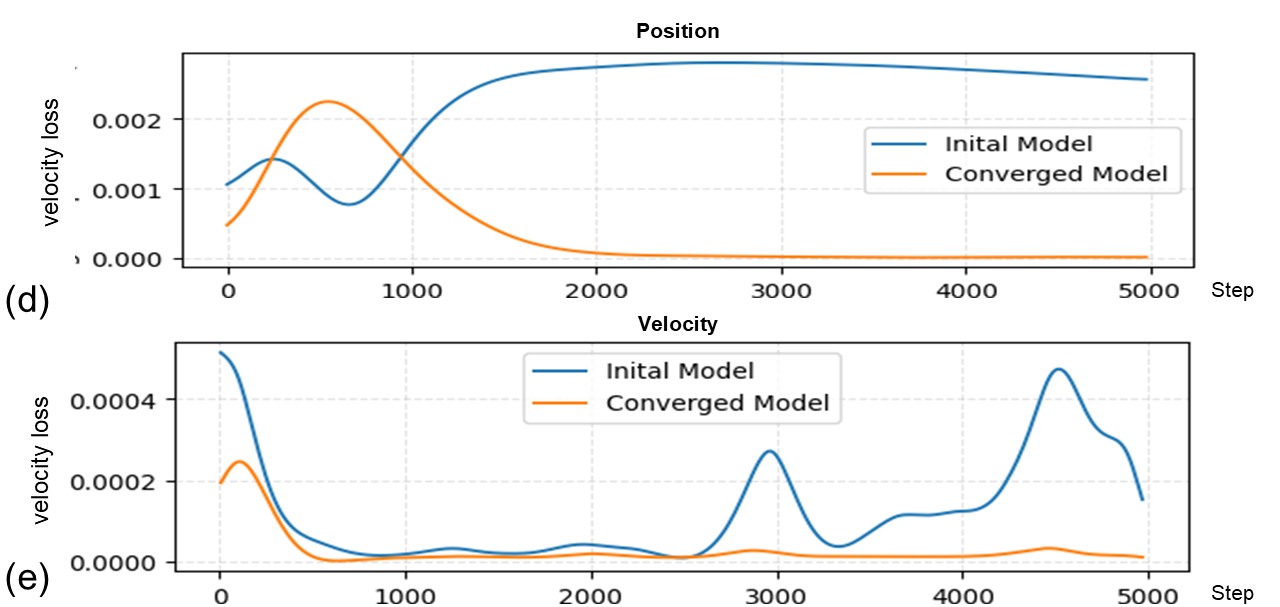}
\caption{(a) Uncurling motion of the physical robot with motion-capture markers illustrating the trajectory over time. (b) Simulated uncurling motion after parameter identification. (c) Simulated uncurling motion using initial, unoptimized parameters. (d) Positional error between the optimised and unoptimized models. (e) Velocity error between the optimized and unoptimized models.}
\label{uncurl}
\vspace{-2mm}
\end{figure}

The qualitative results of our experiments in Table~\ref{table1} are shown in Fig.~\ref{uncurl} (a)–(c). The quantitative results are presented in Fig.~\ref{uncurl} (d) and (e), where a Butterworth low-pass filter is applied for improved interpretability. We define the internal shape error \(E_{\text{internal}}\) as:


\begin{align}
E_{\text{internal}} = \frac{1}{(N - 1) T} \sum_{t=1}^{T} \sum_{i=2}^{N} 
\Big| & \left\| \mathbf{P}_1^{\text{sim}}(t) - \mathbf{P}_i^{\text{sim}}(t) \right\| \notag \\
     & - \left\| \mathbf{P}_1^{\text{real}}(t) - \mathbf{P}_i^{\text{real}}(t) \right\| \Big|
\end{align}
where $N$ is the number of segments in the robot, $T$ is number of frames in the dynamic process.${p}_i^{\text{sim}}(t)$ is the 3D position of segment \(i\) in simulation at frame \(t\).${p}_i^{\text{real}}(t)$ refers to 3D position of segment \(i\) in real-world data at frame \(t\).

The initial model before optimization exhibits an internal error of $8.98~cm$ over $437$ frames, whereas the optimized model after parameter identification achieves an error of $1.58~cm$ over the same number of frames. This corresponds to a reduction from $17\%$ to $3\%$ when normalized by the total length of the soft robot ($50.7~cm$). The fine-tuning process improves model performance by less than 2\%, thereby validating the assumptions derived from the robot’s geometric scaling.


\subsection{Grasping via Augmented Reality Platform}\label{ex-sub3}
\begin{figure*}[htbp]
\centering
\includegraphics[width=0.9\textwidth]{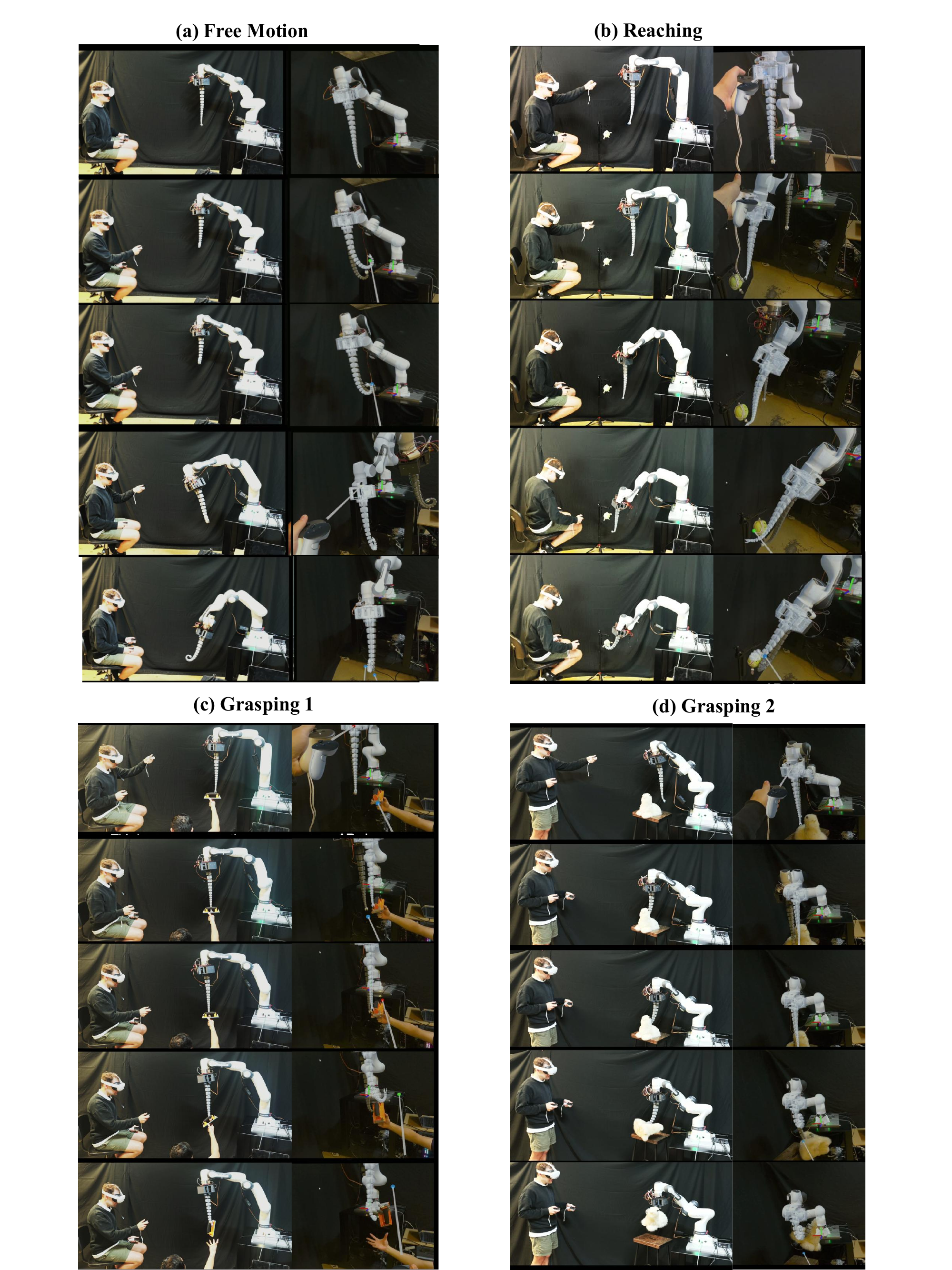} 
\caption{Supplementary videos provide additional details of the demonstration. (a) The user moves the virtual robot freely in 3D space, while the physical robot follows with a designed delay. (b) The user teleoperates the robot to reach a hanging target point. (c) The user teleoperates the robot to grasp an object being handed over in mid-air. (d) The user teleoperates the robot to grasp a toy placed on a chair.}
\label{real_demo}
\end{figure*}

We conduct a series of experiments to demonstrate the teleoperation capabilities of our AR framework. A detailed explanation of each experiment is summarized in Table~\ref{table2}. We also show a series of keyframes of our experiments in Fig.~\ref{real_demo}. We refer readers to our supplementary video to view each experiment.

\begin{table*}[t]
\caption{Human–Robot Interaction Experiments}
\renewcommand\arraystretch{1.5}
        \setlength\tabcolsep{6pt}
\label{table2}
\centering
\begin{tabular}{p{0.3cm}p{2.0cm}p{9.0cm}p{4.2cm}}
\hline
\textbf{Exp.} & \textbf{Objective} & \textbf{Implementation} & \textbf{Success Criteria}  \\
\hline
\textbf{1} & Object Reaching & Reach a static object by using the rigid robot to position and orient the soft robot and by curling the soft robot to bring its tip into contact with the target. & Successful contact between the soft robot's tip and the target object. \\
\hline
\textbf{2} & Object Following & Follow a dynamic trajectory by coordinating rigid-arm positioning and orientation with soft-robot curling and uncurling. & Successful trajectory completion without interruption. \\
\hline
\textbf{3} & Object Grasping & Grasp an object either placed on a flat surface or suspended in mid-air. & Object securely grasped and lifted without slipping or dropping. \\
\hline
\textbf{4} & Object Handling & Grasp an object from one surface, move it in mid-air, and release it onto another surface. & Object grasped, transferred, and released onto the target surface. \\
\hline
\end{tabular}
\vspace{-2.5mm}
\end{table*}

\section{DISCUSSION AND LIMITATIONS}

\subsection{Parameter Identification}

The proposed simulation-centered parameter identification framework effectively aligns simulated and real robot behavior by jointly estimating static, dynamic, and controller parameters. It significantly reduces internal shape error and enables stable teleoperation within the AR framework, providing a practical and reproducible real-to-sim calibration pipeline without requiring analytical continuum models.

However, the simulated controller is a simplified abstraction of the physical system, and unmodeled effects may be implicitly absorbed into the optimized parameters. As a result, the identified parameters are behaviorally consistent but not strictly physically interpretable, limiting their direct use in analytical model-based control approaches such as model predictive control or observer-based methods. Furthermore, the current evaluation focuses primarily on internal error metrics and qualitative demonstrations, and future work should incorporate broader quantitative benchmarks and higher-fidelity physics simulations to improve generalizability.

For applications demanding higher sim-to-real fidelity, such as reinforcement learning, trajectory optimization, or contact-rich manipulation, further sim-to-real adaptation and model refinement beyond the initial real-to-sim identification will be necessary.



\subsection{AR-based Human Robot Interaction}


The proposed AR-based HRI framework introduces a novel target specification and a real-time teleoperation concept for hybrid rigid–soft robots, demonstrating feasibility across trajectory-following, reaching, and grasping tasks. The integration of simulation, AR visualization, and physical execution establishes a coherent interaction pipeline suitable for exploratory manipulation scenarios.

However, a comprehensive quantitative assessment incorporating additional task scenarios, a larger user cohort, and standardized performance metrics is required to rigorously evaluate usability, intuitiveness, and cognitive workload. 




\section{CONCLUSION AND FUTURE WORK}


In this work, we presented a simulation-centered parameter identification framework for a cable-driven soft continuum arm, enabling the estimation of static, dynamic, and controller parameters in MuJoCo. Building on this calibrated simulation, we developed a novel augmented-reality-based human–robot interaction framework for a hybrid rigid–soft robot that enables users to define end-effector targets via a head-mounted display. Our qualitative evaluation demonstrates the feasibility and practical functionality of the integrated real-to-sim and AR teleoperation pipeline. Future work will focus on conducting comprehensive quantitative assessments to systematically evaluate usability, intuitiveness, and cognitive workload. 






\section*{ACKNOWLEDGMENTS}
This research is supported by the National Research Foundation, Prime Ministers Office, Singapore under its Campus for Research Excellence and Technological Enterprise (CREATE) programme. The Mens, Manus, and Machina (M3S) is an interdisciplinary research group of the Singapore MIT Alliance for Research and Technology (SMART) centre. This research is also supported by A*STAR, Singapore, through the Italy-Singapore collaborative project ``DESTRO - Dextrous, strong yet soft robots" (C.L.).

\bibliographystyle{IEEEtran}
\bibliography{root}

@article{montero2024mastering,
  title={Mastering Contact-rich Tasks by Combining Soft and Rigid Robotics with Imitation Learning},
  author={Montero, Mariano Ram{\'\i}rez and Shahabi, Ebrahim and Franzese, Giovanni and Kober, Jens and Mazzolai, Barbara and Della Santina, Cosimo},
  journal={arXiv preprint arXiv:2410.07787},
  year={2024}
}

@article{guan2025control,
  title={Control the Soft Robot Arm with its Physical Twin},
  author={Guan, Qinghua and Cheng, Hung Hon and Dai, Benhui and Hughes, Josie},
  journal={arXiv preprint arXiv:2503.17227},
  year={2025}
}

@article{wu2010overview,
  title={An overview of dynamic parameter identification of robots},
  author={Wu, Jun and Wang, Jinsong and You, Zheng},
  journal={Robotics and computer-integrated manufacturing},
  volume={26},
  number={5},
  pages={414--419},
  year={2010},
  publisher={Elsevier}
}

@inproceedings{khosla1985parameter,
  title={Parameter identification of robot dynamics},
  author={Khosla, Pradeep K and Kanade, Takeo},
  booktitle={1985 24th IEEE conference on decision and control},
  pages={1754--1760},
  year={1985},
  organization={IEEE}
}

@article{zhang2025novel,
  title={A Novel Parameter Estimation Method for Pneumatic Soft Hand Control Applying Logarithmic Decrement for Pseudo-Rigid Body Modeling},
  author={Zhang, Haiyun and Heung, Kelvin HoLam and Naquila, Gabrielle J and Hingwe, Ashwin and Deshpande, Ashish D},
  journal={Advanced Intelligent Systems},
  volume={7},
  number={3},
  pages={2400637},
  year={2025},
  publisher={Wiley Online Library}
}

@article{ouyang2018design,
  title={Design of an interactive control system for a multisection continuum robot},
  author={Ouyang, Bo and Liu, Yunhui and Tam, Hon-Yuen and Sun, Dong},
  journal={IEEE/ASME Transactions on Mechatronics},
  volume={23},
  number={5},
  pages={2379--2389},
  year={2018},
  publisher={IEEE}
}

@article{lai2024gesture,
  title={Gesture-based steering framework for redundant soft robots},
  author={Lai, Jiewen and Lu, Bo and Huang, Kaicheng and Chu, Henry K},
  journal={IEEE/ASME transactions on mechatronics},
  volume={29},
  number={6},
  pages={4651--4663},
  year={2024},
  publisher={IEEE}
}

@inproceedings{stroppa2020human,
  title={Human interface for teleoperated object manipulation with a soft growing robot},
  author={Stroppa, Fabio and Luo, Ming and Yoshida, Kyle and Coad, Margaret M and Blumenschein, Laura H and Okamura, Allison M},
  booktitle={2020 IEEE International Conference on Robotics and Automation (ICRA)},
  pages={726--732},
  year={2020},
  organization={IEEE}
}

@inproceedings{thakur2024tetherless,
  title={A tetherless soft robotic wearable haptic human machine interface for robot teleoperation},
  author={Thakur, Shilpa and Armas, Nathalia Diaz and Adegite, Joseph and Pandey, Ritwik and Mead, Joey and Rao, Pratap M and Onal, Cagdas D},
  booktitle={2024 IEEE/RSJ International Conference on Intelligent Robots and Systems (IROS)},
  pages={12226--12233},
  year={2024},
  organization={IEEE}
}

@article{huang2025grasping,
  title={Grasping by spiraling: reproducing elephant movements with rigid-soft robot synergy},
  author={Huang, Huishi and Wang, Haozhe and Fang, Chongyu and Yan, Mingge and Xu, Ruochen and Zhang, Yiyuan and Wang, Zhanchi and Ying, Fengkang and Liu, Jun and Laschi, Cecilia and others},
  journal={npj Robotics},
  volume={3},
  number={1},
  pages={18},
  year={2025},
  publisher={Nature Publishing Group UK London}
}

@article{dall2025towards,
  title={Towards an intuitive industrial teaching interface for collaborative robots: gamepad teleoperation vs. kinesthetic teaching},
  author={Dall’Alba, Diego and Boriero, Fabrizio},
  journal={The International Journal of Advanced Manufacturing Technology},
  pages={1--18},
  year={2025},
  publisher={Springer}
}

@article{avgousti2016medical,
  title={Medical telerobotic systems: current status and future trends},
  author={Avgousti, Sotiris and Christoforou, Eftychios G and Panayides, Andreas S and Voskarides, Sotos and Novales, Cyril and Nouaille, Laurence and Pattichis, Constantinos S and Vieyres, Pierre},
  journal={Biomedical engineering online},
  volume={15},
  number={1},
  pages={96},
  year={2016},
  publisher={Springer}
}

@article{ajoudani2012tele,
  title={Tele-impedance: Teleoperation with impedance regulation using a body--machine interface},
  author={Ajoudani, Arash and Tsagarakis, Nikos and Bicchi, Antonio},
  journal={The International Journal of Robotics Research},
  volume={31},
  number={13},
  pages={1642--1656},
  year={2012},
  publisher={SAGE Publications Sage UK: London, England}
}

@article{uthai2025haptics,
  title={Haptics-based robot teleoperation for soft object manipulation},
  author={Uthai, Thanakon and Zhou, Tianyu and Ye, Yang and You, Hengxu and Du, Jing},
  journal={Journal of Construction Engineering and Management},
  volume={151},
  number={5},
  pages={04025028},
  year={2025},
  publisher={American Society of Civil Engineers}
}

@inproceedings{walker2019robot,
  title={Robot teleoperation with augmented reality virtual surrogates},
  author={Walker, Michael E and Hedayati, Hooman and Szafir, Daniel},
  booktitle={2019 14th ACM/IEEE International Conference on Human-Robot Interaction (HRI)},
  pages={202--210},
  year={2019},
  organization={IEEE}
}

@article{pan2021augmented,
  title={Augmented reality-based robot teleoperation system using RGB-D imaging and attitude teaching device},
  author={Pan, Yong and Chen, Chengjun and Li, Dongnian and Zhao, Zhengxu and Hong, Jun},
  journal={Robotics and Computer-Integrated Manufacturing},
  volume={71},
  pages={102167},
  year={2021},
  publisher={Elsevier}
}

@article{wang2024spirobs,
  title={SpiRobs: Logarithmic spiral-shaped robots for versatile grasping across scales},
  author={Wang, Zhanchi and Freris, Nikolaos M and Wei, Xi},
  journal={Device},
  year={2024},
  publisher={Elsevier}
}

@inproceedings{bruder2019nonlinear,
  title={Nonlinear system identification of soft robot dynamics using koopman operator theory},
  author={Bruder, Daniel and Remy, C David and Vasudevan, Ram},
  booktitle={2019 International Conference on Robotics and Automation (ICRA)},
  pages={6244--6250},
  year={2019},
  organization={IEEE}
}

@article{giorgio2017hybrid,
  title={Hybrid parameter identification of a multi-modal underwater soft robot},
  author={Giorgio-Serchi, Francesco and Arienti, Andrea and Corucci, Francesco and Giorelli, Michele and Laschi, Cecilia},
  journal={Bioinspiration \& biomimetics},
  volume={12},
  number={2},
  pages={025007},
  year={2017},
  publisher={IOP Publishing}
}

@article{wang2019parameter,
  title={Parameter identification and model-based nonlinear robust control of fluidic soft bending actuators},
  author={Wang, Tao and Zhang, Yunce and Chen, Zheng and Zhu, Shiqiang},
  journal={IEEE/ASME transactions on mechatronics},
  volume={24},
  number={3},
  pages={1346--1355},
  year={2019},
  publisher={IEEE}
}

@article{wang2017soft,
  title={Soft gripper dynamics using a line-segment model with an optimization-based parameter identification method},
  author={Wang, Zhongkui and Hirai, Shinichi},
  journal={IEEE Robotics and Automation Letters},
  volume={2},
  number={2},
  pages={624--631},
  year={2017},
  publisher={IEEE}
}

@InProceedings{HeLiu2021,
  author    = {Yanhao He and Steven Liu},
  booktitle = {2021 9th International Conference on Control, Mechatronics and Automation (ICCMA2021)},
  title     = {Analytical Inverse Kinematics for {F}ranka {E}mika {P}anda -- a Geometrical Solver for 7-{DOF} Manipulators with Unconventional Design},
  year      = {2021},
  month     = nov,
  publisher = {{IEEE}},
  doi       = {10.1109/ICCMA54375.2021.9646185},
}

@inproceedings{todorov2012mujoco,
  title={MuJoCo: A physics engine for model-based control},
  author={Todorov, Emanuel and Erez, Tom and Tassa, Yuval},
  booktitle={2012 IEEE/RSJ International Conference on Intelligent Robots and Systems},
  pages={5026--5033},
  year={2012},
  organization={IEEE},
  doi={10.1109/IROS.2012.6386109}
}

\end{document}